%% file: ArxivPreprintSubmission2027.tex
\documentclass[letterpaper]{article} 
\usepackage[preprint]{aaai2027}  
\usepackage[hyphens]{url}  
\usepackage{graphicx} 
\usepackage{docmute} 
\usepackage{natbib}  
\usepackage{caption} 
\usepackage{algorithm}
\usepackage{algorithmic}
\usepackage{booktabs}
\usepackage{makecell}
\usepackage{multirow}
\usepackage{bbding}
\usepackage{colortbl}
\usepackage{amsmath}

\makeatletter
\let\varsdocmaketitle\maketitle
\let\varsdocatmaketitle\@maketitle
\makeatother

\title{VaRS-Doc: Interpretation-Aware Variant Representations \\ via Latent Self-Probing for Visual Document Retrieval}
\author{
    Haocheng Wang\textsuperscript{\rm 1},
    Tongkun Guan\textsuperscript{\rm 1},
    Wei Shen\textsuperscript{\rm 1}\corresponding,
    Xiaokang Yang\textsuperscript{\rm 1}
}
\affiliations{
    \textsuperscript{\rm 1}MoE Key Lab of Artificial Intelligence, AI Institute, School of Computer Science,\\
    Shanghai Jiao Tong University, China\\
    \{wanghaocheng2023, wei.shen\}@sjtu.edu.cn
}

\begin{document}

\maketitle

\begin{abstract}
Visual document retrieval has recently become increasingly important in applications such as enterprise search, scientific literature discovery, and retrieval-augmented generation. These applications depend on efficiently identifying query-relevant pages across large collections of visually rich documents. Existing methods commonly adopt late-interaction architectures that encode and index documents offline to enable scalable and low-latency online retrieval. Despite its efficiency, this paradigm requires each document to be encoded into a fixed representation before the query is known. However, the same content in a visual document may induce different interpretations depending on the query intent, which a fixed representation struggles to capture. Yet postponing document encoding until the query arrives would incur prohibitive online retrieval latency.
To address this gap, we propose VaRS-Doc, a visual document retrieval framework that diversifies document representations by enabling the model to actively explore variant latent interpretations during document encoding, while preserving efficient late-interaction retrieval in which each query adaptively selects the best-fit representation. We further introduce a two-stage training strategy that encourages the
model to capture complementary semantic interpretations and prevents it from falling back to train a single dominant representation. Experiments on visual document retrieval benchmarks show that VaRS-Doc achieves state-of-the-art retrieval performance, offering a practical solution to the mismatch between query-agnostic document encoding and query-specific retrieval needs. Code is available at \url{https://github.com/bokufa/VaRS-Doc}.
\end{abstract}

\section{Introduction}

\begin{figure}[!t]
    \centering
    \includegraphics[width=\columnwidth]{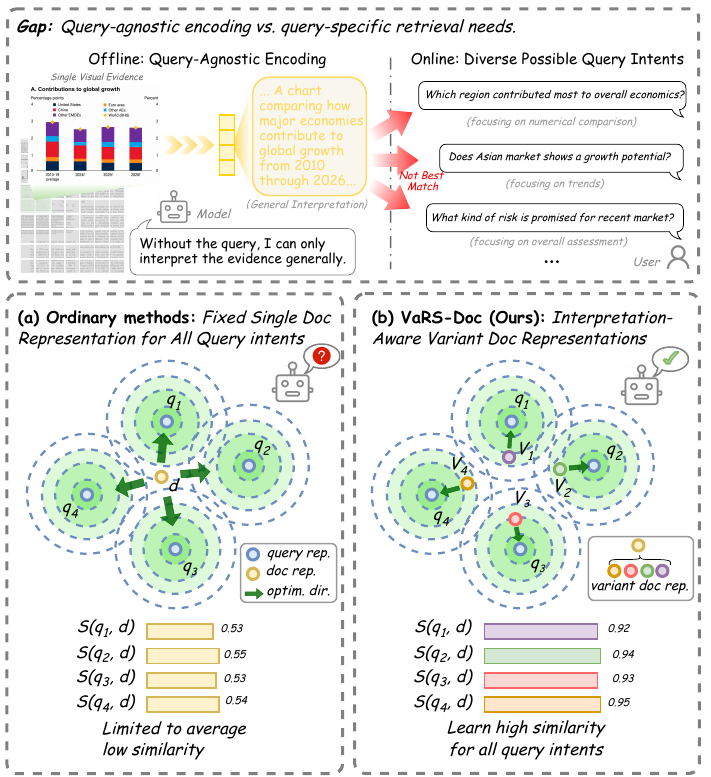}
    \caption{Motivation of VaRS-Doc. Scalable retrieval requires query-agnostic document encoding, creating a gap against query-specific retrieval needs. (a) Existing methods align queries with diverse intents with single document representation, limiting its alignment. (b) VaRS-Doc instead provides interpretation-aware variants, allowing queries to be better aligned with its best-matched variant.}    
    \label{fig:teaser}
    \vspace{-20pt}
\end{figure}

Visual document retrieval has become important for accessing information in large collections of visually rich documents, such as scientific papers, business reports, slides, manuals, and scanned PDFs~\cite{lewis2020rag}. A retrieval model is therefore expected to search over document images and identify pages that are relevant to a natural language query. Unlike conventional text retrieval, visual document retrieval must handle heterogeneous evidence, including paragraphs, tables, figures, and layout structures. This makes the task both practically important and semantically challenging.

Early visual document retrieval methods apply text retrieval models to OCR-extracted content, but inevitably lose important visual cues such as layout, typography, tables, and figures~\cite{robertson2009bm25,chen2024m3embedding,zhang2025ocrhindersrag}. Recent methods such as ColPali and VisRAG~\cite{faysse2025colpali,yu2025visrag,alayrac2022flamingo,li2022blip} therefore turn to multimodal large language models (MLLMs) to perform fine-grained visual-textual semantic matching. However, applying MLLMs to large-scale retrieval introduces a severe efficiency challenge: the corpus may contain millions of pages, whereas user queries arrive in real time. To make retrieval scalable, most practical methods decouple query and document encoding by adopting a Late-Interaction~\cite{khattab2020colbert,faysse2025colpali} architecture. Documents are encoded offline and stored in an index, while each incoming query is encoded online and compared with the indexed documents.

However, this efficiency comes with a fundamental constraint: once a document has been indexed, its representation is fixed and cannot adapt to the intent of a later query. Therefore, index models can only be trained to interpret the document in a general way for all possible queries.  Meanwhile, the same visual evidence can support different semantic readings. For example, the chart shown in Fig.~\ref{fig:teaser} may be examined to compare regional contributions, identify growth trends in specific markets, or assess recent market risks, depending on the query intent. In such cases, relevance depends not only on whether the evidence is present, but also on which semantic interpretation of that evidence is activated by the query. Therefore, a general-purpose document representation would fail to match the specific interpretation needed for a given query, limiting retrieval performance, as shown in Fig.~\ref{fig:teaser}(a).
Some methods~\cite{qwen3vlembedding} instead postpone document encoding until the query arrives and cross-encode the query and document to predict a relevance score for the pair, allowing document evidence to be interpreted under the specific query intent. However, this cross encoding process must be repeated for all the documents at query time and the resulting representations can not be reused for future queries, making them impractical for full-range large-scale retrieval.

This reveals a key gap: document representations must be computed offline before the queries are known for scalable retrieval, making it impossible for the model to adaptively encode specific interpretations of the same visual evidence for different query intents.
This gap motivates us to ask whether part of the benefit of query-conditioned interpretation can be obtained without sacrificing the scalability of offline indexing. To this end, we propose \textbf{VaRS-Doc} (\emph{Interpretation-Aware \textbf{Va}riant \textbf{R}epresentations via Latent \textbf{S}elf-Probing for Visual \textbf{Doc}ument Retrieval}), a retrieval framework that encourages the model to internally explore complementary interpretations of the same document evidence before any user query is observed. The key idea behind \emph{latent self-probing} is that the model itself probes a document from different semantic perspectives during encoding. Specifically, VaRS-Doc encodes each page together with a set of learnable \emph{latent interpretation probing tokens}. These tokens first read the shared document context without affecting the encoding of the document tokens, becoming document-conditioned probes. Each probe then guides a separate branch of encoding process to differentiate the document states into a variant interpretation representation of the document. The resulting variants therefore capture complementary interpretations of the same visual evidence. This \textit{Shared-Contextualization with Branched-Encoding} design first establishes a rich, document-specific context shared by all variants and then allows each branch to develop a distinct interpretation on top of it, while also avoiding repeating the full document encoding for every variant. At retrieval time, the incoming query scores
a document by selecting the most relevant variant. In this way, as shown in Fig.~\ref{fig:teaser}(b), VaRS-Doc enables each query to adaptively activate the specific best-fit reading of a document, while preserving the efficiency of late-interaction retrieval.

Training these representations from random initialization can cause the model to rely on a single dominant branch for most queries, leaving the others poorly trained. We address this with a two-step training strategy: we first learn a reliable retriever with one representation and then expand it to multiple representations, ensuring that every branch begins multi-variant training with competitive retrieval ability. The second step then encourages different branches to win on different training examples with a clear best match for each query. This keeps all representations useful and allows them to capture complementary readings of a document.

Finally, our contributions are summarized as follows:
\begin{itemize}
    \item We propose \textbf{VaRS-Doc}, a novel retrieval framework that actively reads and encodes documents in complementary interpretations, enabling different queries to adaptively select best-fit readings of documents, while preserving the efficiency of late-interaction architecture.
    
    \item We develop a training strategy that ensures different variants are effectively used and learn complementary interpretations, preventing the model from relying on a single dominant branch.

    \item Experimental results show that VaRS-Doc achieves state-of-the-art retrieval performance, while further analyses confirm that the learned variants are effectively used and capture complementary interpretations.
\end{itemize}

\begin{figure*}[t]
    \centering
    \includegraphics[width=\textwidth]{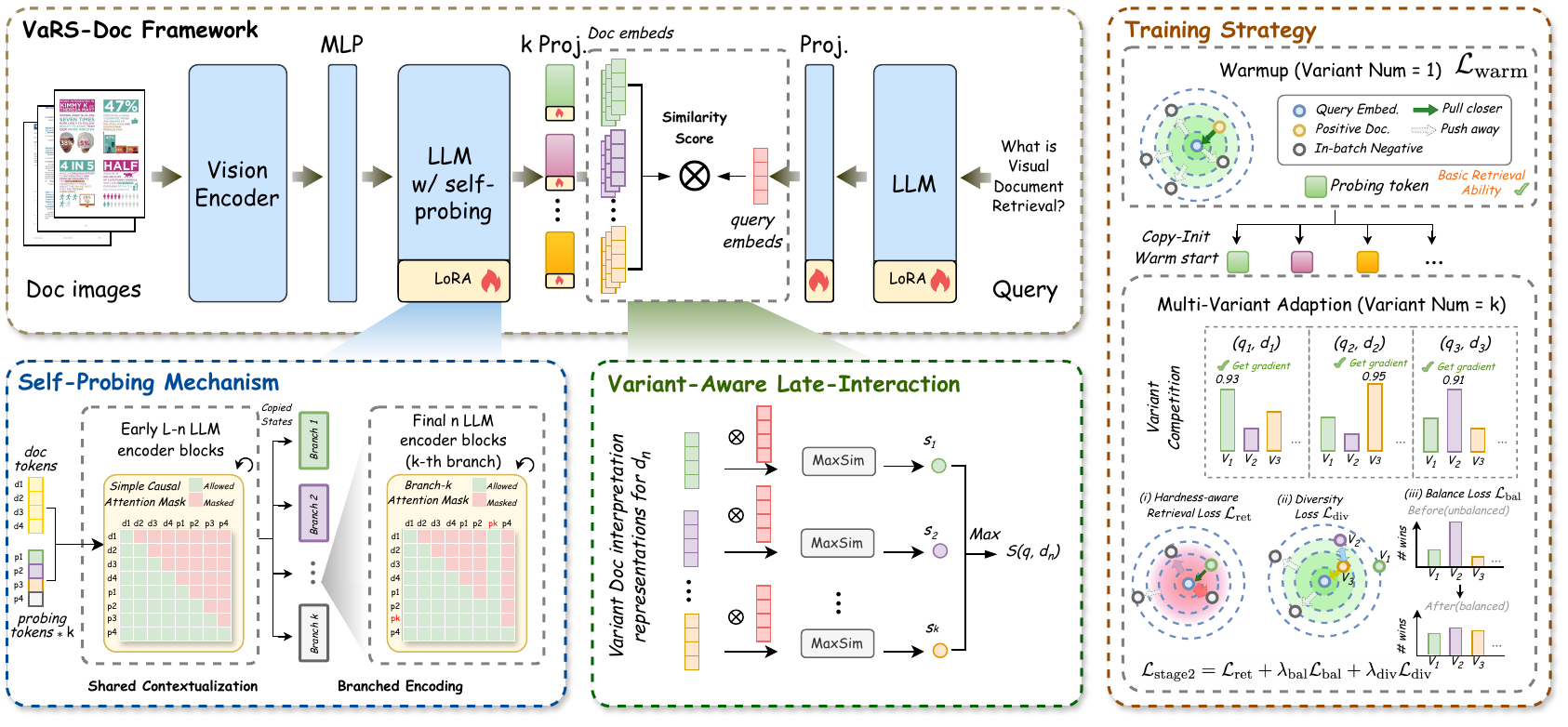}
    \caption{Overview of VaRS-Doc. During offline indexing, latent interpretation probing
    tokens silently read the document context in Shared Contextualization and then guide
    separate branches in Branched Encoding to produce complementary variant representations.
    At retrieval time, variant-aware late interaction selects the best-matched precomputed
    variant for each query. Two-stage variant optimization prevents branch collapse and
    encourages complementary specialization.}
    \label{fig:framework}
\end{figure*}

\section{Related Work}

\paragraph{Multimodal Large Language Models for Document Understanding.}
Multimodal large language models have made rapid progress in understanding visually rich documents~\cite{fu2025mllmsurvey,alayrac2022flamingo,li2022blip,tschannen2025siglip2}. Early document understanding methods, such as LayoutLM~\cite{xu2020layoutlm} and LayoutLMv2~\cite{xu2021layoutlmv2}, incorporate OCR tokens and layout information to model document structure. Later OCR-free models, including Donut~\cite{kim2022donut} and Pix2Struct~\cite{lee2023pix2struct}, directly process document images for structured visual understanding~\cite{guan2025tokenfd}. Recent MLLMs, such as LLaVA~\cite{liu2023llava}, PaliGemma~\cite{beyer2024paligemma}, Qwen-VL~\cite{bai2023qwenvl}, and InternVL~\cite{chen2024internvl,zhu2025internvl3}, further strengthen visual-textual alignment and reasoning over text-rich images. These models provide strong foundations for visual document retrieval, where both textual content and visual layout need to be captured.

\paragraph{Visual Document Retrieval.}
Visual document retrieval aims to retrieve relevant pages from image-based document collections using natural-language queries. Traditional approaches extract OCR text and apply text retrieval models such as BM25~\cite{robertson2009bm25,zhang2025ocrhindersrag}, but often lose important layout and visual information. Recent MLLM-based retrievers~\cite{jiang2024e5v,jiang2025vlm2vec,gunther2025jinav4,nussbaum2024nomicvision,nomic2025multimodal,guan2026lightstar,zeng2021beyond,zeng2023beyond,zeng2024focus,qin2025clip}, including ColPali~\cite{faysse2025colpali}, ColQwen~\cite{faysse2025colpali, bai2025qwen25vl}, and VisRAG~\cite{yu2025visrag}, directly encode document images and adopt late-interaction architectures, where document representations are computed offline and matched with independently encoded queries~\cite{khattab2020colbert}. Qwen3-Reranker~\cite{qwen3vlembedding} instead postpones document encoding until the query arrives and cross-encodes each query--document pair to directly predict its relevance score. Although this enables document evidence to be interpreted according to the query intent, the repeated online document encoding makes it unsuitable for first-stage retrieval at scale. Our approach, VaRS-Doc retains efficient offline indexing while preparing multiple document interpretations from which each query can select.

\section{Method}

\noindent\textbf{Problem Formulation.}
Given a natural language query $q$ and a large collection of visual document pages
$\mathcal{D}=\{d_1,d_2,\ldots,d_N\}$, visual document retrieval aims to rank all pages
according to their relevance to the query. The retrieval objective is formulated as
\begin{equation}
    [d_{\pi_1}, d_{\pi_2}, \ldots, d_{\pi_N}]
    =
    \operatorname*{argsort}_{d_i \in \mathcal{D}}
    S(q,d_i),
\end{equation}
where $S(q,d_i)$ denotes the relevance score between query $q$ and document page
$d_i$.

Most practical large-scale retrieval systems adopt a late-interaction architecture, where
the query and document are encoded independently,
\begin{equation}
    Z_Q=f_Q(q),\
    Z_D=f_D(d), \
    S(q,d)=h(Z_Q,Z_D).
\end{equation}
where $f_Q(\cdot)$ and $f_D(\cdot)$ denote the query and document encoders,
$Z_Q$ and $Z_D$ are the corresponding embeddings, and
$h(\cdot,\cdot)$ is a lightweight similarity function.
Document embeddings can therefore be precomputed and indexed offline.
While this design enables efficient first-stage retrieval, the document representation is
constructed before the query is observed and therefore cannot adapt to different query
interpretations. Meanwhile, cross-encoding methods that jointly encode the query and document can adaptively interpret the document under the specific query intent, but they require expensive online computation for each query-document pair and cannot reuse document representations for future queries.

This motivates us to seek an approach that encodes each document for diverse
possible interpretations rather than compress it into a fixed representation during offline indexing, while retaining the efficiency of late interaction.

\subsection{Method Overview}

Following the above motivation, we propose \textbf{VaRS-Doc}, a novel visual document retrieval framework that actively reads and encodes documents from complementary interpretive perspectives. Instead of forcing a document page into a single representation before knowing how it will be queried, VaRS-Doc encourages the model to internally explore multiple latent interpretations of the same document evidence during offline indexing. These interpretations serve as prepared semantic views that can later be matched against different query intents.

Specifically, as illustrated in Fig.~\ref{fig:framework}, VaRS-Doc prepares each visual document for
possible interpretations through \textit{latent self-probing}. Specifically, given a
document page, we insert a set of learnable latent interpretation probing tokens into
the document encoder. These tokens read the document content and serve as internal
anchors for different potential query intents. To produce such interpretations
efficiently, VaRS-Doc adopts a \textit{Shared-Contextualization with Branched-Encoding} architecture: the
early encoder blocks are shared to encode the document content once, while the final
encoder blocks branch into complementary variant streams guided by probing
tokens. The resulting document variants are precomputed and indexed offline, and are
later matched with the independently encoded query under the \textit{variant-aware late-interaction} paradigm.
To train these prepared interpretation variants to be effectively used and complementary, we further introduce a two-stage variant
optimization strategy, which first learns a strong single-representation retriever and then
expands it for multi-interpretation training.

\subsection{Variant Interpretation Encoding via Self-Probing}
\paragraph{Latent Interpretation Probing Tokens}

In offline document indexing, the document encoder must represent a page before any
future query is known. Compressing the page into one fixed representation can erase
distinctions among interpretations relevant to different query intents. We therefore let
the model explore possible readings through learnable latent probes rather than
predefined natural-language questions.

Specifically, given a page $d$ represented by $M$ document tokens
$X_D=\{x_1,\ldots,x_M\}$, we introduce $K$ learnable \textit{latent interpretation
probing tokens} $\mathcal{P}=\{p_1,\ldots,p_K\}$. These tokens are concatenated after
the document tokens, acting as latent encoding
instructions. They first read the page context, and then each probing token provides a distinct
probing instruction for encoding the same document under a different interpretation.

The learnable probing tokens are optimized directly through retrieval supervision,
allowing them to discover complementary, task-relevant reading directions and adapt them
to each document after reading its context. This avoids manually specifying possible
reading aspects and reduces sensitivity to the wording and length of explicit
natural-language prompts. As a result, VaRS-Doc can prepare documents for
diverse future query intents beyond a manually defined set of interpretations.

\paragraph{Shared Contextualization with Branched Encoding}
To leverage latent probing tokens for interpreting each document page from diverse perspectives, we divide document encoding into two stages: shared context reading and probe-guided branching. The probing tokens first silently read the document context without affecting the encoding of the document tokens; the resulting document-aware probes then guide different branches toward complementary interpretations of the same document.

\noindent\textit{i) Shared Semantic Contextualization.}
Let the document LLM encoder contain $L$ transformer blocks, with the final $n$
blocks used for branched encoding. The first $L-n$ blocks perform shared
contextualization. Given the document token sequence $X_D$ and probing
tokens $\mathcal{P}$, we append the probing tokens to the end of the document token sequence, forming the input 
\begin{equation} 
U^{(0)} = [X_D;\mathcal{P}] = [x_1,\ldots,x_M,p_1,\ldots,p_K]. 
\end{equation} 
This ordering allows us to naturally reuse the causal attention mask in the LLM blocks. Since document tokens appear before the probing tokens, they cannot attend to the later probes. Meanwhile, each probing token can attend to the preceding document tokens. The shared contextualization stage is therefore computed as 
\begin{equation}
U^{(\ell)}
=
F_{\ell}\bigl(U^{(\ell-1)}; A_{\mathrm{causal}}\bigr),
\qquad \ell=1,\ldots,L-n,
\end{equation}
where $F_{\ell}$ denotes the $\ell$-th transformer block. After the first $L-n$ blocks, the
output hidden sequence can be decomposed as
\begin{equation}
U^{(L-n)} = [H_D^{(L-n)}; H_{\mathcal{P}}^{(L-n)}].
\end{equation}
where $H_D^{(L-n)}$ contains the $M$ shared document states, while
$H_{\mathcal{P}}^{(L-n)}$ contains the $K$ document-conditioned probe states. The states in
$H_{\mathcal{P}}^{(L-n)}$ are no longer generic learnable tokens, but page-specific latent probes
that guide the subsequent variant interpretation encoding.

\noindent\textit{ii) Probe-Guided Branched Encoding.}
The final $n$ blocks perform probe-guided branching. Specifically, we first duplicate the shared hidden sequence into $K$
variant streams:
\begin{equation}
    V_k^{(L-n)}
    =
    U^{(L-n)}
    ,\qquad k=1,\ldots,K .
\end{equation}
Thus, each branch still contains all document states and all document-conditioned
probe states.

Branch specialization is controlled by a branch-specific binary visibility mask $A_{\mathrm{br}}^{(k)}$, where $A(i,j)=1$ permits the token at position $i$ to attend to that at position $j$. Since $[H_D^{(L-n)};H_{\mathcal{P}}^{(L-n)}]$ contains $M$ document states followed by $K$ probe states, the $k$-th probe is located at position $M+k$. Branch $k$ extends the causal mask $A_{\mathrm{causal}}$ by allowing every document state to attend to this probe:\begin{equation}
A_{\mathrm{br}}^{(k)}(i,j) = A_{\mathrm{causal}}(i,j) \lor \mathbf{1}\bigl[1 \leq i \leq M,\ j = M+k\bigr], 
\end{equation} 
where $\lor$ denotes element-wise logical OR and $\mathbf{1}[\cdot]$ is the indicator function. Thus, branch $k$ preserves the original causal visibility while exposing only its corresponding probe state in $H_{\mathcal{P}}^{(L-n)}$ to the document states.

These final $n$ transformer blocks are then applied to each branch with its own attention
mask:
\begin{equation}
    V_k^{(\ell)}
    =
    F_{\ell}\bigl(V_k^{(\ell-1)}; A_{\mathrm{br}}^{(k)}\bigr),
    \quad
    \ell=L-n+1,\ldots,L.
\end{equation}
After the final block, the hidden states of the $k$-th branch are grouped as
\begin{equation}
    V_k^{(L)}
    =
    [H_{D,k}^{(L)}; H_{\mathcal{P},k}^{(L)}],
\end{equation}
where $H_{D,k}^{(L)}$ and $H_{\mathcal{P},k}^{(L)}$ contain the final document and probe states,
respectively. The document states are then projected into the retrieval embedding
space after discarding the probe states:
\begin{equation}
    Z_D^{(k)}
    =
    \operatorname{Proj}_k\bigl(H_{D,k}^{(L)}\bigr).
\end{equation}
Therefore, probe tokens do not directly enter the final query matching.
They influence retrieval only by guiding the document token representations during the
branched encoding.

This masking-based branching preserves the original token order and positional
embeddings of the document tokens. Each branch starts from the same shared document
states, but gains access to a different document-conditioned probe state, allowing the
final document embeddings to specialize toward different latent aspects of the same
page.

\subsection{Variant-Aware Late Interaction}

After offline indexing, each document page $d$ is represented by a set of variant
embeddings
\begin{equation}
    \mathcal{Z}_D
    =
    \{Z_D^{(1)}, Z_D^{(2)}, \ldots, Z_D^{(K)}\},
\end{equation}
where $Z_D^{(k)}$ denotes the $k$-th probe-guided document variant. At retrieval time,
the query is encoded independently as
\begin{equation}
    Z_Q = f_Q(q).
\end{equation}
where $f_Q$ processes the query with the same MLLM backbone in document encoding
and projects its output hidden states into the shared retrieval space.
For each variant $k\in\{1,\ldots,K\}$, we compute its late-interaction MaxSim score directly as~\cite{khattab2020colbert}
\begin{equation}
s_k(q,d)
=
\sum_{u=1}^{|Z_Q|}
\max_{1\leq v\leq |Z_D^{(k)}|}
\left\langle z_{q,u},z_{d,v}^{(k)}\right\rangle ,
\end{equation}
where $z_{q,u}$ is the $u$-th query-token embedding and
$z_{d,v}^{(k)}$ is the $v$-th document-token embedding in variant $k$.

The final relevance score is obtained by selecting the best-matched variant:
\begin{equation}
    S(q,d)
    =
    \max_{k=1}^{K} s_k(q,d).
\end{equation}
This variant-aware late-interaction score evaluates each branch independently and lets
the query select the best-matched one, keeping the outputs and learning signals of different branches separate and
allowing each branch to develop a coherent document-level interpretation.

\subsection{Two-stage Variant Optimization}

Training multiple document variants from random initialization can be unstable, since different
branches may start to compete before the model has learned a reliable retrieval space and quickly fall into relying on one variant who dominates most query-document pairs while the remaining variants receive little effective supervision.
We therefore adopt a two-stage optimization strategy to address this.

\noindent\textbf{Single-Variant Warmup.}
We first train a non-branched retriever with one probe token and one document
variant. This stage follows the standard late-interaction training pipeline and learns a
stable query-document embedding space. After convergence, the resulting
checkpoint serves as a strong initialization point for multi-variant training. When this
initialization is copied to multiple variants, every variant enters training as a strong
competitor. The subsequent training can therefore focus on differentiating the variants
into complementary interpretations rather than learning basic retrieval ability from
scratch.

Given a mini-batch of $B$ positive query-document pairs $\{(q_i,d_i)\}_{i=1}^{B}$, the optimization objective is as below:
\begin{equation} 
\mathcal{L}_{\mathrm{warm}} = \frac{1}{B} \sum_{i=1}^{B} \log\left(1+\exp\left(\max_{j\ne i}S(q_i,d_j)-S(q_i,d_i)\right)\right). 
\end{equation}

\noindent\textbf{Multi-Variant Adaptation.}
We then expand the single-variant model into a $K$-variant model. Specifically, the
learned probe token and the
projection head are copied into $K$ branch instances. The following variant-aware
objective then breaks this symmetry and drives branch specialization.

\noindent\textit{i) Hardness-Aware Retrieval Loss.}
We redesign the standard InfoNCE loss~\cite{oord2018cpc} into a \textit{hardness-aware InfoNCE loss} to train model to focus on resolving
challenging negatives. Specifically, we increase the contribution of negative documents according to their current similarity scores, so that training focuses more on confusing query-document pairs while still preserving the efficiency of in-batch contrastive learning.
\begin{equation}
    \mathcal{L}_{\mathrm{ret}}
    =
    -\frac{1}{B}
    \sum_{i=1}^{B}
    \log
    \frac{\exp(S_{ii})}
    {\exp(S_{ii})
    +
    \sum_{j\neq i} w_{ij}\exp(S_{ij})},
\end{equation}
where $w_{ij}$ is hardness weight computed from detached score: \begin{equation} w_{ij} = \operatorname{sg}\left(\exp(S_{ij})\right), \qquad j\neq i, \end{equation} where $\operatorname{sg}(\cdot)$ denotes the stop-gradient operation.

\noindent\textit{ii) Batch-Level Balance Loss.}
To further avoid long-term domination by a single variant, we introduce a batch-level balance
loss. For each positive query-document pair $(q_i,d_i)$, we use $s_{ii}^{(k)}$ to denote
the score between query $q_i$ and the $k$-th variant of its positive document. We convert these scores into a probability-like usage distribution over variants: a higher score means that the corresponding variant is more responsible for matching this sample. The average usage of the $k$-th variant within a mini-batch $B$ is computed as 
\begin{equation} 
\bar{a}_k = \frac{1}{B} \sum_{i=1}^{B} \frac{\exp(s_{ii}^{(k)})} {\sum_{k'=1}^{K}\exp(s_{ii}^{(k')})} . 
\end{equation}
The balance loss encourages the batch-level variant usage to be close to a uniform
distribution:
\begin{equation}
    \mathcal{L}_{\mathrm{bal}}
    =
    \sum_{k=1}^{K}
    \left(\bar{a}_k-\frac{1}{K}\right)^2 .
\end{equation}
This loss encourages variant branch usage across a batch, preventing one
branch from dominating all training samples.

\begin{table*}[!t]
    \centering
    \renewcommand{\arraystretch}{1.05}
    \caption{
        Performance comparison on the ViDoRe V2 and V3 benchmarks.
        Results are reported using NDCG@10 (\%).
        $^{\dagger}$ denotes our direct backbone baseline.
        The best and second-best results are shown in
        \textbf{bold} and \underline{underlined}, respectively.
    }
    \label{tab:main_results}

    \resizebox{\textwidth}{!}{
    \begin{tabular}{l c cccc|c| cccccccc|c| c}
        \toprule

        \multirow{2}{*}{\textbf{Model}}
        & \multirow{2}{*}{\textbf{Size}}
        & \multicolumn{5}{c}{\textbf{ViDoRe V2}}
        & \multicolumn{9}{c}{\textbf{ViDoRe V3}}
        & \multirow{2}{*}{\textbf{All}} \\[-0.2ex]

        \cmidrule(lr){3-7}
        \cmidrule(lr){8-16}

        & &
        \textbf{Bio.}
        & \textbf{Econ.}
        & \textbf{ESG-S}
        & \textbf{ESG-H}
        & \textbf{Avg.}
        & \textbf{HR}
        & \textbf{Fin.-EN}
        & \textbf{Ind.}
        & \textbf{Pharm.}
        & \textbf{CS}
        & \textbf{Energy}
        & \textbf{Physics}
        & \textbf{Fin.-FR}
        & \textbf{Avg.}
        & \\[-0.2ex]

        \midrule
        
        \textbf{\emph{Vision-Language Contrastive Models}}\\

        CLIP~\cite{radford2021clip}
        & 149M
        & 11.5
        & 6.2
        & 7.4
        & 14.1
        & 9.8
        & 4.6
        & 1.4
        & 2.1
        & 14.3
        & 6.1
        & 4.7
        & 8.2
        & 0.8
        & 5.3
        & 6.8 \\

        SigLIP~\cite{zhai2023siglip}
        & 877M
        & 30.7
        & 18.2
        & 39.4
        & 50.1
        & 34.6
        & 15.0
        & 16.0
        & 8.9
        & 29.0
        & 28.6
        & 16.0
        & 17.8
        & 6.6
        & 17.2
        & 23.0 \\

        \midrule
        
        \textbf{\emph{MLLM-Based Retrieval Models}}
         \\

        ColPali~\cite{faysse2025colpali}
        & 3B
        & 58.8
        & 48.4
        & 57.4
        & 62.4
        & 56.7
        & 44.8
        & 34.4
        & 35.6
        & 53.1
        & 65.3
        & 46.9
        & 41.7
        & 21.8
        & 43.0
        & 47.5 \\

        eager-embed~\cite{eagerworks2025eagerembed}
        & 4B
        & \underline{65.5}
        & 50.3
        & 57.5
        & 60.6
        & 58.5
        & 49.3
        & 44.5
        & 35.2
        & 60.3
        & 70.9
        & 56.3
        & 46.5
        & 31.0
        & 49.3
        & 52.3 \\

        SauerkrautLM-ColQwen3~\cite{vago2025sauerkraut}
        & 2B
        & 62.5
        & \underline{53.8}
        & 57.3
        & \underline{72.9}
        & \underline{61.6}
        & 53.0
        & 54.3
        & 44.0
        & 60.6
        & \underline{73.7}
        & 61.2
        & \underline{47.6}
        & 40.2
        & 54.3
        & 56.8 \\

        ColNomic-Embed-Multimodal~\cite{nomic2025multimodal}
        & 3B
        & 65.4
        & 53.6
        & 53.3
        & 60.7
        & 58.2
        & \textbf{57.3}
        & \underline{56.3}
        & \underline{47.4}
        & \underline{61.1}
        & 72.7
        & \textbf{64.5}
        & 47.6
        & \textbf{44.3}
        & \underline{56.4}
        & \underline{57.0} \\

        \midrule

        ColQwen2.5~\cite{faysse2025colpali,bai2025qwen25vl}$^{\dagger}$
        & 3B
        & 63.1
        & 52.8
        & \underline{62.4}
        & 68.2
        & \underline{61.6}
        & 51.2
        & 52.3
        & 41.3
        & 57.9
        & 72.3
        & 59.5
        & 45.9
        & 39.1
        & 52.5
        & 55.5 \\

        \rowcolor{gray!20}
        \textbf{VaRS-Doc \emph{(Ours)}}
        & 3B
        & \textbf{66.8}
        & \textbf{59.9}
        & \textbf{64.2}
        & \textbf{75.2}
        & \textbf{66.5}
        & \underline{54.3}
        & \textbf{57.2}
        & \textbf{49.0}
        & \textbf{62.8}
        & \textbf{75.5}
        & \underline{63.4}
        & \textbf{48.2}
        & \underline{42.9}
        & \textbf{56.7}
        & \textbf{60.0} \\

        \bottomrule
    \end{tabular}
    }
\end{table*}

\noindent\textit{iii) Batch-Level Diversity Loss.}
Complementary to batch-level balancing, we apply a top margin diversity loss to enlarge the winner margin.
For the positive pair
$(q_i,d_i)$, the best variant score is exactly the document-level score
$S_{ii}$. We define the second-best variant score as $R_{ii}$.
The diversity loss is then defined as
\begin{equation}
    \mathcal{L}_{\mathrm{div}}
    =
    \frac{1}{B}
    \sum_{i=1}^{B}
    \max\bigl(0, \gamma - (S_{ii}-R_{ii})\bigr),
\end{equation}
where $\gamma$ is a small target margin. This loss encourages the best-matched variant to
outperform the second-best variant by at least $\gamma$, avoiding ambiguous cases where variants produce nearly identical scores for the same query.

Finally, the objective of the second stage is as below: 
\begin{equation} 
\mathcal{L}_{\mathrm{stage2}} = \mathcal{L}_{\mathrm{ret}} + \lambda_{\mathrm{bal}}\mathcal{L}_{\mathrm{bal}} + \lambda_{\mathrm{div}}\mathcal{L}_{\mathrm{div}}, 
\end{equation} 
where $\lambda_{\mathrm{bal}}$ and $\lambda_{\mathrm{div}}$ control the strengths of balance loss and diversity loss, respectively. 

The balance loss and diversity loss together create meaningful competition in which different branches win for different queries, driving the variants of the same document to specialize toward complementary interpretations.

In both stages, we apply Low-Rank Adaptation (LoRA)~\cite{hu2022lora} to the transformer layers of the language-model
backbone, latent probing tokens and the output projection
layers. This training design keeps the
adaptation lightweight and stable.

\section{Experiments}

\paragraph{Implementation Details.}
We use ColQwen2.5~\cite{faysse2025colpali, bai2025qwen25vl} as backbone, with $K=5$ document variants and $n=4$ branched transformer blocks. We first warm up the single-variant retriever on the training dataset~\cite{faysse2025colpali} for three epochs with a learning rate of $5\times10^{-4}$, followed by two epochs of multi-variant adaptation with a learning rate of $5\times10^{-5}$. We use four NVIDIA A800 GPUs with a batch size of 32 per GPU. LoRA is applied with rank $r=32$ and scaling factor $\alpha=32$. We set $\lambda_{\mathrm{bal}}=0.1$, $\lambda_{\mathrm{div}}=0.05$, and $\gamma=0.05$.
Additional details and hyperparameter analyses are provided in Appendix.

\paragraph{Datasets.}
We evaluate VaRS-Doc on the ViDoRe V2 and V3 benchmarks~\cite{mace2025vidorev2,loison2026vidorev3,muennighoff2023mteb}, which together cover 12 retrieval datasets across diverse domains and document formats. Document pages are provided directly as images. Following prior work, we report NDCG@10 as the metric.

\subsection{Main Results}

As shown in Table~\ref{tab:main_results}, VaRS-Doc achieves state-of-the-art performance on both ViDoRe V2 and V3, obtaining average NDCG@10 scores of 66.5 and 56.7, respectively, and an overall average of 60.0. Since VaRS-Doc is directly built upon the ColQwen2.5 retrieval framework with the same MLLM backbone, ColQwen2.5 serves as the most relevant comparison for isolating the effectiveness of our proposed variant representations. VaRS-Doc consistently outperforms ColQwen2.5 on all 12 datasets, improving the V2 and V3 averages by 4.9 and 4.2 points, respectively, and the overall average by 4.5 points. Particularly large gains are observed on Economics Reports (+7.1), Industrial Documents (+7.7), and human-labeled ESG Reports (+7.0), demonstrating that query-adaptive variant representations provide more effective coverage of the diverse semantic interpretations contained in visual documents.

\paragraph{Interpretation Complementarity Analysis.}
We randomly sample $N=100$ pages in ViDoRe V2 and select one localized evidence region from each.
Given only this region, an MLLM generates two distinct query intents and two
paraphrases per intent. We manually verify intent distinctness, paraphrase
equivalence, and that every query is answerable from the selected region.
For page $d_i$, let $q_{i,a,m}$ denote formulation $m$ of intent $a$, where
$a,m\in\{1,2\}$. The selected variant and comparison probabilities are
\begin{equation}
\begin{aligned}
v_{i,a,m}
&= \operatorname*{argmax}_{k\in\{1,\ldots,K\}}
s_k(q_{i,a,m},d_i), \\
P_{\mathrm{same}}
&= \frac{1}{2N}\sum_{i=1}^{N}\sum_{a=1}^{2}
\mathbf{1}[v_{i,a,1}=v_{i,a,2}], \\
P_{\mathrm{cross}}
&= \frac{1}{4N}\sum_{i=1}^{N}\sum_{m,n=1}^{2}
\mathbf{1}[v_{i,1,m}=v_{i,2,n}].
\end{aligned}
\end{equation}
Here, $P_{\mathrm{same}}$ and $P_{\mathrm{cross}}$ measure the same-variant
probability for same- and different-intent query pairs, respectively.

\begin{table}[!t]
  \centering
  \caption{Probability that query pairs select the same document variant (\%).}
  \label{tab:interpretation_complementarity}
  \renewcommand{\arraystretch}{1.15}
  \setlength{\tabcolsep}{7pt}
  \resizebox{0.45\textwidth}{!}
  {\begin{tabular}{lcc}
    \toprule
    \textbf{Model}
    & \textbf{\begin{tabular}{@{}c@{}}Same Intent ($P_{\mathrm{same}}$)\end{tabular}}
    & \textbf{\begin{tabular}{@{}c@{}}Different Intents ($P_{\mathrm{cross}}$)\end{tabular}} \\
    \midrule
    VaRS-Doc & 77.6 & 30.8 \\
    \bottomrule
  \end{tabular}}
\end{table}

As shown in Table~\ref{tab:interpretation_complementarity},
same-intent paraphrases select the same variant in 77.6\% of cases, compared
with 30.8\% for different intents. This result shows that the variants remain
stable across paraphrases while capturing complementary semantics of
the same evidence under different query intents.

 \begin{table}[t]
  \centering
  \caption{Efficiency Analysis. All results are measured
  using the same ViDoRe V2 corpus, hardware, and retrieval configuration.
  Values in parentheses are relative to baseline.}
  \label{tab:efficiency}
  \resizebox{0.45\textwidth}{!}{
  \begin{tabular}{lccc}
  \toprule
  Method & Index Size / Doc & Index Time / Doc & Latency / Query \\
  \midrule
  Baseline ($K=1$)

  & 1.48 MiB (1.00$\times$)
  & 201.85 ms (1.00$\times$)
  & 22.38 ms (1.00$\times$) \\

  VaRS-Doc ($K=3$)

  & 4.44 MiB (3.00$\times$)
  & 203.26 ms (1.01$\times$)
  & 26.40 ms (1.18$\times$) \\

  VaRS-Doc ($K=5$)

  & 7.41 MiB (5.00$\times$)
  & 230.72 ms (1.14$\times$)
  & 27.15 ms (1.21$\times$) \\
  \bottomrule
  \end{tabular}
  }
  \end{table}

\paragraph{Efficiency Analysis.}
Table~\ref{tab:efficiency} shows that the index size grows linearly
with the number of variants as expected, since each variant stores a full document
representation. In contrast, index time increases by only $1.01\times$
and $1.14\times$, as document encoding also includes the vision encoder
and the shared LLM blocks, while only the final few LLM blocks are branched;
the branched design therefore shows a modest increase in index time.
Online, the query is encoded only once, and additional variants require only
lightweight MaxSim scoring and score maximization rather than extra MLLM forward
passes. Since late-interaction similarity computation is inherently fast, this
additional scoring increases query latency only modestly, from 22.38 ms to 26.40
ms and 27.15 ms. Such a small absolute overhead poses no substantial barrier to
practical online deployment, yielding a favorable trade-off between retrieval
effectiveness and efficiency.

\subsection{Ablation Studies}
\label{sec:ablation}

We conduct ablation studies to examine the contributions of the variant
representation design and the proposed
optimization strategy. Unless otherwise specified, all experiments use the
default settings with five document variants, and four branched
Transformer blocks ($n=4$). We report the averaged NDCG@10 on ViDoRe V2.

\begin{table}[!t]
    \centering
    \caption{
        Ablation of the variant representation design on ViDoRe V2.
        We report the average NDCG@10 (\%).
    }
    \label{tab:ablation_variant_design}
    \renewcommand{\arraystretch}{1.15}
    \setlength{\tabcolsep}{8pt}
    \resizebox{0.45\textwidth}{!}{
    \begin{tabular}{l|cccc}
        \toprule
        \textbf{Setting}
        &
        \textbf{\begin{tabular}{@{}c@{}}Latent\\Probing\end{tabular}}
        &
        \textbf{\begin{tabular}{@{}c@{}}Branched\\Encoding.\end{tabular}}
        &
        \textbf{\begin{tabular}{@{}c@{}}Variant\\Projection\end{tabular}}
        &
        \textbf{NDCG@10} \\
        \midrule

        Baseline (ColQwen)
        & \XSolidBrush
        & \XSolidBrush
        & \XSolidBrush
        & 61.6 \\

        \quad + variant projection
        & \XSolidBrush
        & \XSolidBrush
        & \Checkmark
        & 61.8 \\

        \quad + probing token
        & \Checkmark
        & \XSolidBrush
        & \XSolidBrush
        & 62.1 \\

        \quad + branched encoding
        & \Checkmark
        & \Checkmark
        & \XSolidBrush
        & 65.8 \\

        \rowcolor{gray!20}
        \textbf{Full Model}
        & \Checkmark
        & \Checkmark
        & \Checkmark
        & \textbf{66.5} \\

        \bottomrule
    \end{tabular}
    }

\end{table}

\paragraph{Variant Representation Design.}
Table~\ref{tab:ablation_variant_design} evaluates the proposed variant
representation design, together with a projection-only control that tests
whether its gains can be attributed simply to increased output capacity.
Adding multiple variant-specific projection heads without latent probing or
branched encoding improves NDCG@10 only from 61.6 to 61.8. This marginal gain
shows that the full-model improvement
does not primarily come from increased output capacity: projection alone is
insufficient to produce distinct interpretations, and the main semantic
differentiation must occur within the MLLM encoder. Introducing latent self-probing alone brings only
a modest improvement, increasing the average NDCG@10 from 61.6 to 62.1. In
contrast, enabling branched contextualization substantially raises the score to
65.8, indicating that the main benefit comes from allowing different branches to
develop distinct document interpretations rather than merely appending probing
tokens. Variant-specific projection heads provide a further improvement to 66.5.
Overall, the full design outperforms the single-representation baseline by 4.9
points, with branched encoding accounting for the majority of the gain.

\paragraph{Optimization Strategy.}
Table~\ref{tab:ablation_optimization} evaluates the main components of our
two-stage optimization. Removing the single-variant warm-up causes the largest
performance degradation, reducing NDCG@10 from 66.5 to 63.7. More importantly,
the maximum branch usage increases from 26.2\% to 94.2\%, showing that direct
multi-variant training causes nearly all samples to be assigned to a single
dominant branch. Replacing hardness-weighted InfoNCE with the standard objective
also lowers NDCG@10 to 64.8 and increases both the tie rate and branch
concentration, confirming that emphasizing informative hard negatives benefits
both retrieval accuracy and variant specialization.
The two auxiliary losses regulate complementary aspects of the variant
assignment. Without $\mathcal{L}_{\mathrm{bal}}$, the maximum branch usage
rises sharply to 92.4\%, indicating severe global branch collapse. Although
this setting exhibits a relatively low tie rate, the result is caused by one
branch consistently dominating rather than by meaningful specialization.
In contrast, removing $\mathcal{L}_{\mathrm{div}}$ increases Tie Rate@0.01
from 18.9\% to 56.8\%, showing that the variants become difficult to
distinguish within individual query--document pairs. The full objective
achieves the best NDCG@10 while maintaining substantially more balanced branch
utilization and clear query-dependent preferences. These results demonstrate
that warm-up and $\mathcal{L}_{\mathrm{bal}}$ prevent global collapse, whereas
$\mathcal{L}_{\mathrm{div}}$ promotes confident within-sample specialization.

\begin{table}[!t]
  \centering
  \caption{
      Ablation of the optimization strategy on ViDoRe V2, averaged over
      four datasets. Tie Rate@0.01 is the fraction of samples with near-tied
      top variants, and Max Branch Usage is the largest branch winning ratio;
      lower is better for both.
  }
  \label{tab:ablation_optimization}
  \renewcommand{\arraystretch}{1.15}
  \setlength{\tabcolsep}{7pt}
  \resizebox{0.47\textwidth}{!}{
  \begin{tabular}{l|ccc}
      \toprule
      \textbf{Setting}
      & \textbf{NDCG@10}
      & \textbf{\begin{tabular}{@{}c@{}}Tie Rate\\@0.01\end{tabular}}
      & \textbf{\begin{tabular}{@{}c@{}}Max Branch\\Usage\end{tabular}} \\
      \midrule

      w/o Warm-up
      & 63.7 & 11.5 & 94.2 \\

      w/o hardness weighting
      & 64.8 & 24.8 & 31.7 \\

      w/o $\mathcal{L}_{\mathrm{bal}}$
      & 65.8 & 11.6 & 92.4 \\

      w/o $\mathcal{L}_{\mathrm{div}}$
      & 65.3 & 56.8 & 27.9 \\

    \rowcolor{gray!20}
      \textbf{Full Model}
      & \textbf{66.5}
      & \textbf{18.9}
      & \textbf{26.2} \\

      \bottomrule
  \end{tabular}
  }
  \vspace{-5pt}
\end{table}

\section{Conclusion}

In this paper, we presented VaRS-Doc, a novel visual document retrieval framework that provides interpretation-aware variant representations for document encoding. VaRS-Doc enables self-probing encoding using a Shared Contextualization with Branched Encoding architecture to generate variant document representations, while variant-aware late interaction allows each query to select the best-matched variant efficiently.
By preparing different readings before queries are observed, VaRS-Doc provides a practical solution to the gap between the query-agnostic document encoding and the query-specific interpretation needs for accurate matching. Comprehensive experiments demonstrate state-of-the-art performance and verify the effectiveness of the proposed components.
We want to emphasize that, beyond simple performance gains, VaRS-Doc introduces a novel retrieval paradigm: preparing a single fixed representation for any possible query is limiting the model. Expressive power of the retrieval model can be further revealed by allowing it to probe more possible interpretations.


\bibliography{aaai2027}

\clearpage
\makeatletter
\let\maketitle\varsdocmaketitle
\let\@maketitle\varsdocatmaketitle
\let\c@aaai@eqfn\@undefined
\let\c@aaai@corrfn\@undefined
\let\titlearea\@undefined
\let\actualheight\@undefined
\makeatother
\title{VaRS-Doc: Interpretation-Aware Variant Representations\\
via Latent Self-Probing for Visual Document Retrieval\\(Supplementary Material)}
\author{}
\affiliations{}
\setcounter{secnumdepth}{2}

\input{SupplementaryMaterial2027.inc}


\end{document}

%% file: SupplementaryMaterial2027.inc
\maketitle

\appendix
\numberwithin{figure}{section}
\numberwithin{table}{section}
\numberwithin{equation}{section}

\section{Implementation and Experimental Details}
\label{sec:supp-details}

\noindent\textbf{Basic Setup.}
VaRS-Doc uses ColQwen2.5 with five document variants and four branched
Transformer blocks. It is trained on four NVIDIA A800 GPUs using a three-epoch
single-variant warm-up followed by two epochs of multi-variant adaptation. The
complete settings are listed in Table~\ref{tab:supp-hyperparameters}.

\begin{center}
    \captionof{table}{Hyperparameter settings used in our experiments.}
    \label{tab:supp-hyperparameters}
    
    \begin{tabular*}{0.8\columnwidth}{@{\hspace{1em}\extracolsep{\fill}}cc@{\hspace{1em}}}
        \toprule
        \textbf{Setting} & \textbf{Value} \\
        \midrule
        Backbone & ColQwen2.5 \\
        Variants $K$ & 5 \\
        Branched $n$ & 4 \\
        GPUs & $4\times$ A800 \\
        Batch/GPU & 32 \\
        LoRA rank $r$ & 32 \\
        LoRA scaling $\alpha$ & 32 \\
        Warm-up epochs & 3 \\
        Warm-up LR & $5\times10^{-4}$ \\
        Adapt. epochs & 2 \\
        Adapt. LR & $5\times10^{-5}$ \\
        $\lambda_{\mathrm{bal}}$ & 0.1 \\
        $\lambda_{\mathrm{div}}$ & 0.05 \\
        Margin $\gamma$ & 0.05 \\
        \bottomrule
    \end{tabular*}
\end{center}

\noindent\textbf{Training Dataset.}
We train VaRS-Doc on the same training dataset used by
ColPali~\cite{faysse2025colpali}. It contains 118,695 English query--page
pairs, combining the training splits of DocVQA, InfoVQA, TAT-DQA, and arXivQA
with synthetic pairs constructed from pages in web-crawled PDFs.

\noindent\textbf{Efficiency Measurement.}
All testing experiments are conducted on a single NVIDIA A800 GPU with a batch
size of 32. Each efficiency measurement is repeated 10 times, and we report the
average result. We measure efficiency on the ViDoRe V2 document corpus under
the same hardware and retrieval configuration. \emph{Index Size / Doc} measures
stored embeddings per page; \emph{Index Time / Doc} measures the time to encode
and store one page; and \emph{Latency / Query} measures online query encoding
and retrieval from the same corpus.

\section{Additional Experimental Results}
\label{sec:supp-experimental-results}

\subsection{Ablation Study}
\label{sec:supp-ablation}

\noindent\textbf{Variant-Aware Late Interaction.}
We compare three strategies for aggregating multiple document variants.
\emph{Mean over Variants} averages the independently computed variant scores,
\emph{Token-Level Fusion} combines all variant tokens before late interaction,
and \emph{Max over Variants} scores each variant separately and retains the
highest score.

\begin{table}[ht]
    \centering
    \caption{Ablation of the variant aggregation strategy on ViDoRe V2. We
    report the average NDCG@10 (\%).}
    \label{tab:supp-ablation-aggregation}
    \setlength{\tabcolsep}{10pt}
    \begin{tabular}{lc}
        \toprule
        \textbf{Aggregation Strategy} & \textbf{NDCG@10} \\
        \midrule
        Mean over Variants & 66.2 \\
        Token-Level Fusion & 66.3 \\
        \rowcolor{gray!20}
        Max over Variants & \textbf{66.5} \\
        \bottomrule
    \end{tabular}
\end{table}

As shown in Table~\ref{tab:supp-ablation-aggregation}, max-over-variants
achieves the best NDCG@10 of 66.5, compared with 66.2 for mean aggregation and
66.3 for token-level fusion. Keeping the variants separate prevents the
evidence captured by one branch from being mixed with or diluted by other
branches, allowing each branch to develop a complete document interpretation.
The query can then directly select the interpretation that best matches its
intent.

\begin{figure}[h]
    \centering
    \begin{minipage}[t]{0.465\linewidth}
        \centering
        \includegraphics[width=\linewidth]{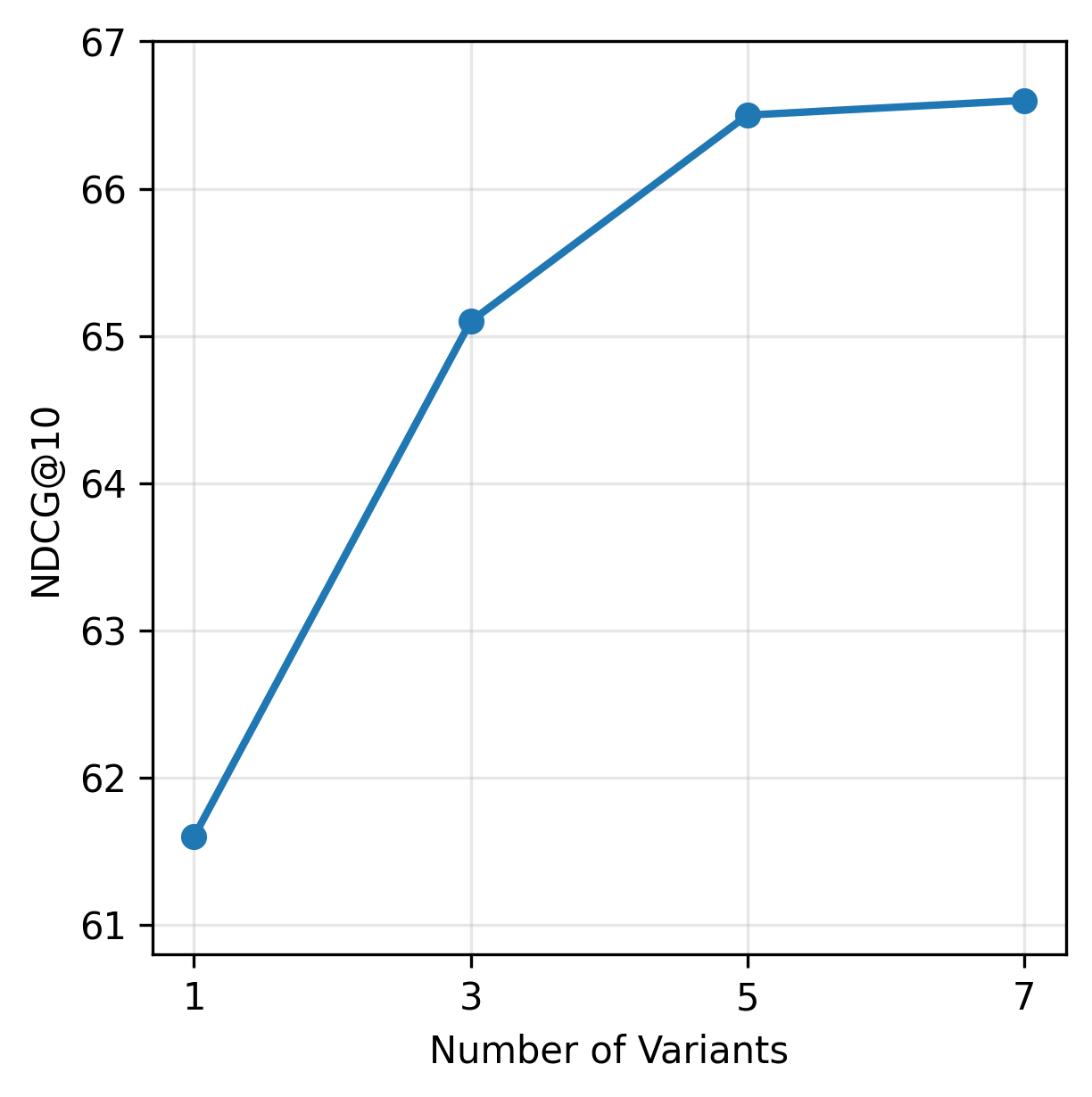}
        \\[-0.8ex]
        \scriptsize (a) Number of variants.
    \end{minipage}
    \hfill
    \begin{minipage}[t]{0.48\linewidth}
        \centering
        \includegraphics[width=\linewidth]{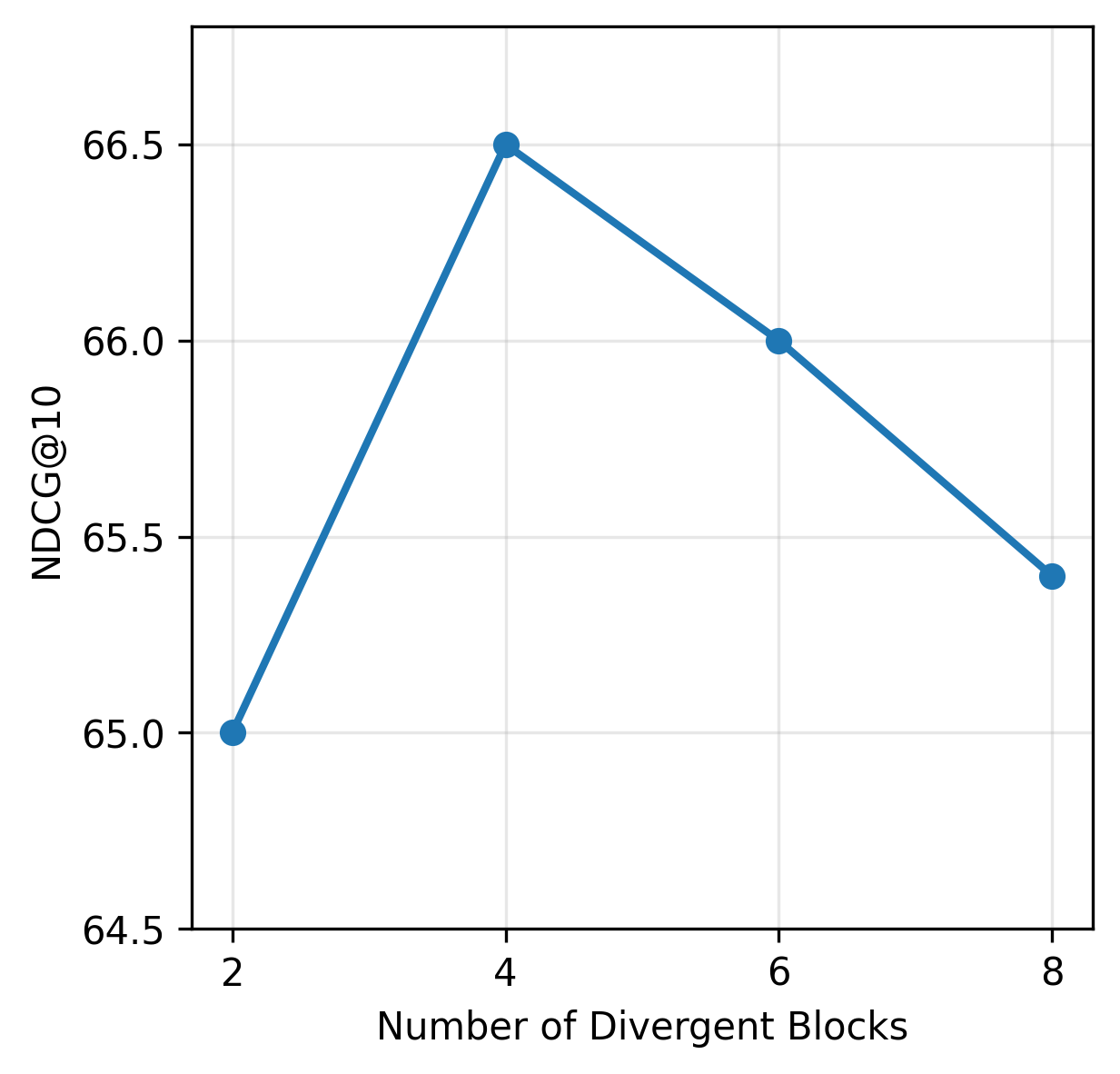}
        \\[-0.8ex]
        \scriptsize (b) Branching depth.
    \end{minipage}
    \caption{Effects of the number of variants and branching depth on ViDoRe
    V2. We report the average NDCG@10 (\%) over its four datasets.}
    \label{fig:supp-variant-scaling}
\end{figure}

\begingroup
\renewcommand{\thefigure}{C.1}
\begin{figure*}[t]
    \centering
    \includegraphics[width=\textwidth]{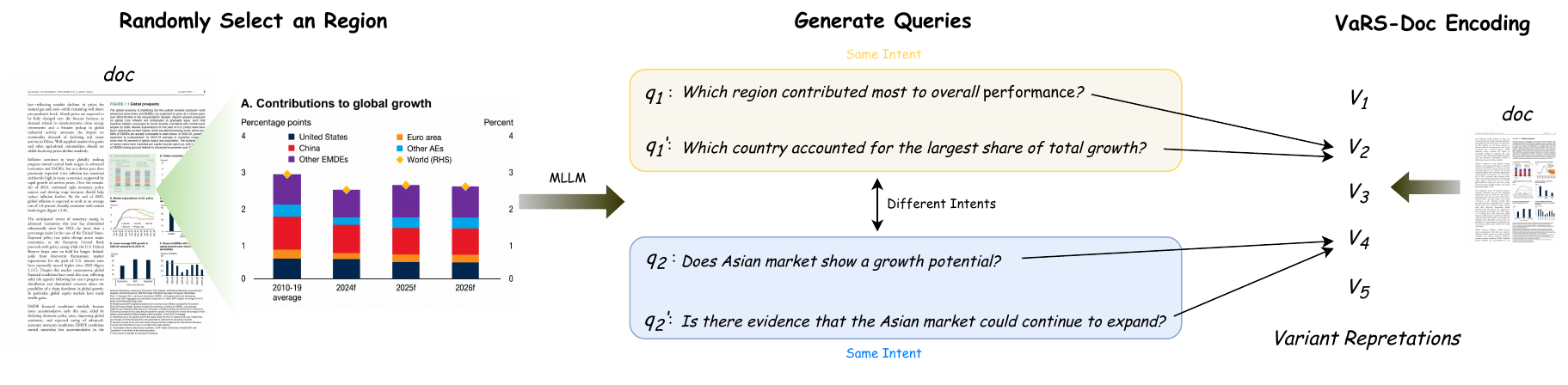}
    \caption{Example of interpretation complementarity analysis. Given a
    localized evidence region, Qwen3.5-9B generates two query intents and one
    meaning-preserving paraphrase for each intent. Same-intent queries select
    the same document variant, whereas queries with different intents select
    different variants.}
    \label{fig:supp-interpretation-sample}
\end{figure*}
\addtocounter{figure}{-1}
\endgroup

\begin{figure}[H]
    \centering
    \begin{minipage}[t]{0.48\linewidth}
        \centering
        \includegraphics[width=\linewidth]{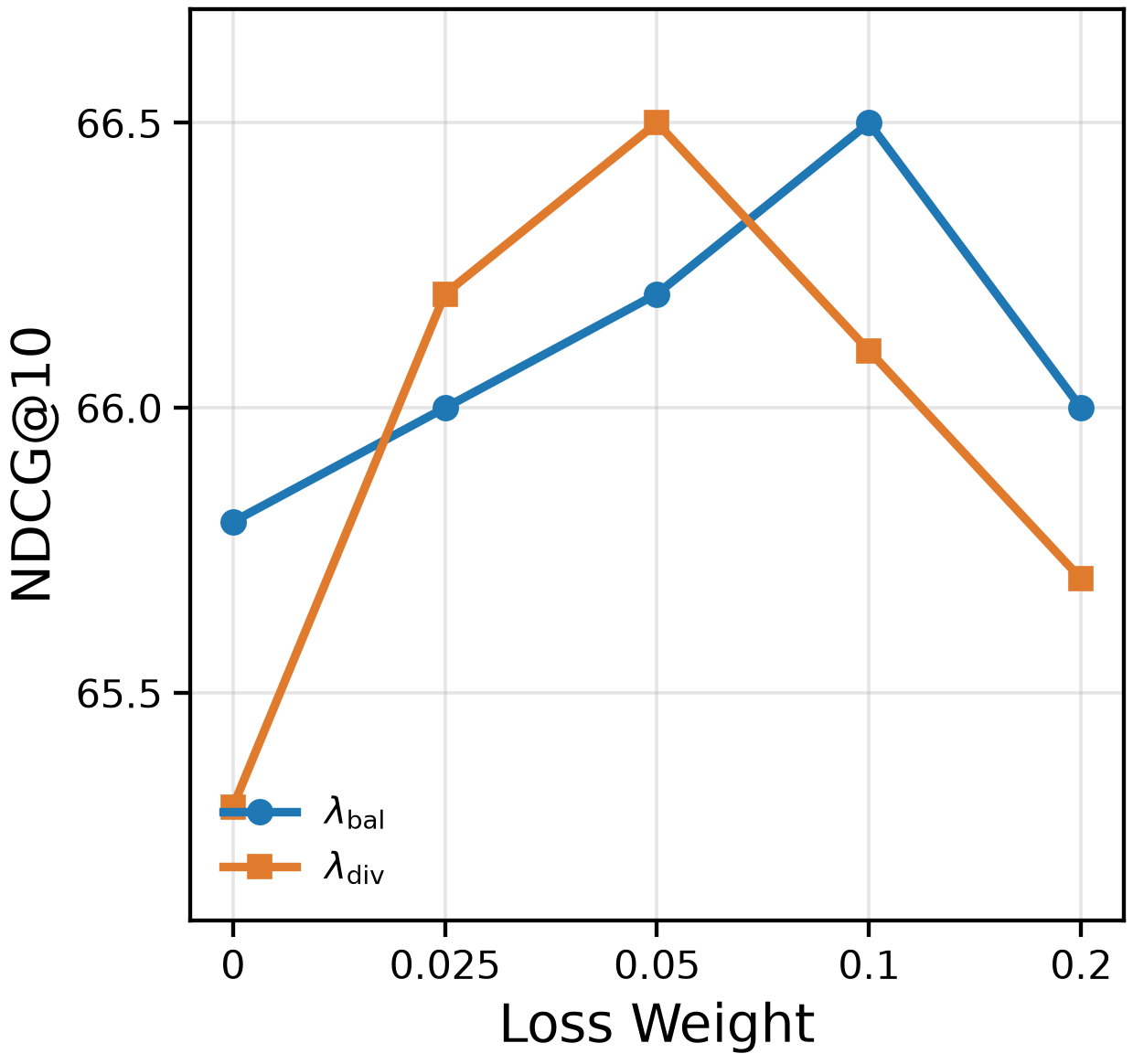}
        \\[-0.8ex]
        \scriptsize (a) Loss weights.
    \end{minipage}
    \hfill
    \begin{minipage}[t]{0.48\linewidth}
        \centering
        \includegraphics[width=\linewidth]{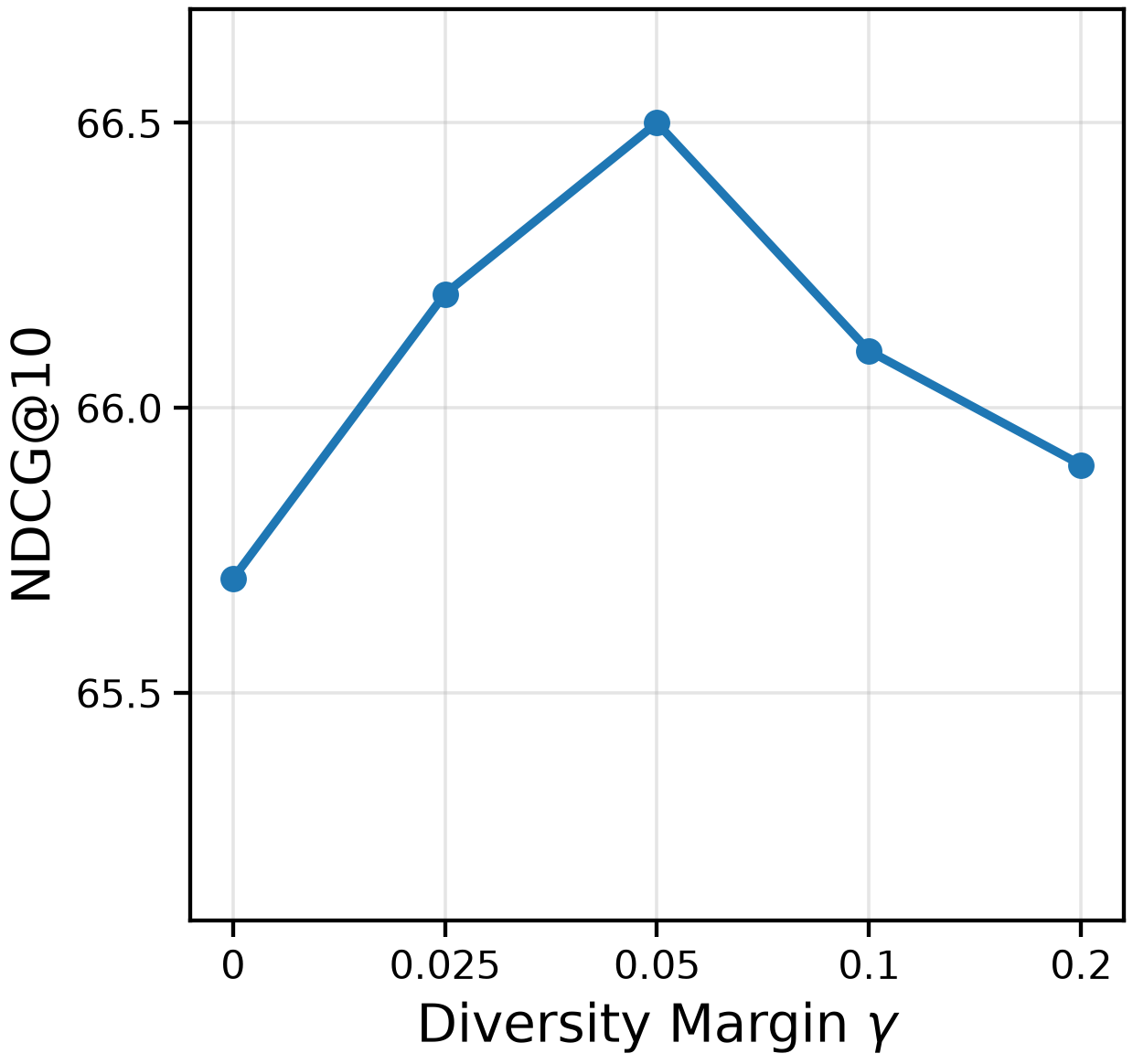}
        \\[-0.8ex]
        \scriptsize (b) Diversity margin $\gamma$.
    \end{minipage}
    \caption{Effects of the loss weights and diversity margin on ViDoRe V2.
    We report the average NDCG@10 (\%) over its four datasets.}
    \label{fig:supp-loss-hyperparameters}
\end{figure}

\subsection{Hyperparameter Analysis}
\label{sec:supp-hyperparameter-analysis}

\noindent\textbf{Number of Variants and Branching Depth.}
We further examine how the number of document variants and the branching depth
affect retrieval performance. As shown in
Figure~\ref{fig:supp-variant-scaling}(a), increasing the number of variants from
one to three improves NDCG@10 from 61.6 to 65.1, demonstrating the benefit of
representing a document through multiple interpretations. Performance further
increases to 66.5 with five variants, whereas increasing the number to seven
yields only a marginal gain. This suggests that five variants provide
sufficient interpretation coverage, while additional variants increase storage
and computational costs with limited benefit.

Figure~\ref{fig:supp-variant-scaling}(b) shows that using two branched
Transformer blocks achieves 65.0 NDCG@10, suggesting insufficient capacity for
the branches to develop differentiated interpretations. Increasing the depth to
four blocks improves performance to 66.5. Branching earlier by using six or
eight blocks instead decreases performance to 66.0 and 65.4, respectively,
because it weakens shared document contextualization. Four branched blocks
therefore provide the best balance between shared document understanding and
interpretation specialization.

\noindent\textbf{Loss Weights and Diversity Margin.}
Figure~\ref{fig:supp-loss-hyperparameters}(a) shows that both regularization
objectives benefit from moderate weights. As discussed in the main paper, the
two losses prevent complementary collapse modes. Removing
$\mathcal{L}_{\mathrm{bal}}$ increases the maximum branch usage to 92.4\%,
indicating global branch collapse in which one variant dominates most samples.
Removing $\mathcal{L}_{\mathrm{div}}$ instead raises Tie Rate@0.01 to 56.8\%,
corresponding to a within-sample collapse toward nearly indistinguishable
variant scores. Increasing $\lambda_{\mathrm{bal}}$
from 0 to 0.1 improves NDCG@10 from 65.8 to 66.5, whereas a larger weight of 0.2
reduces it to 66.0. Similarly, $\lambda_{\mathrm{div}}=0.05$ achieves the best
performance of 66.5, while increasing it to 0.2 decreases NDCG@10 to 65.7,
suggesting that overly strong diversity regularization separates variants at
the expense of retrieval relevance.

Figure~\ref{fig:supp-loss-hyperparameters}(b) further shows that a small
positive diversity margin is important. Performance increases from 65.7 with
$\gamma=0$ to 66.5 with $\gamma=0.05$, but falls to 65.9 when the margin is
increased to 0.2. This indicates that a moderate margin prevents variant
collapse, whereas an excessively large margin forces unnecessary separation.
We therefore use $\lambda_{\mathrm{bal}}=0.1$,
$\lambda_{\mathrm{div}}=0.05$, and $\gamma=0.05$ in all experiments.

\section{Interpretation Complementarity Analysis}
\label{sec:supp-interpretation}

\noindent\textbf{MLLM-Based Query Generation.}
We use Qwen3.5-9B to generate queries. From 100 randomly sampled ViDoRe V2
pages, we select one localized evidence region per page and provide only that
crop to the MLLM. It generates two semantically distinct queries and one
meaning-preserving paraphrase for each. We manually verify intent distinctness,
paraphrase equivalence, and answerability from the selected evidence. We then
evaluate the four queries against the original page and record the
highest-scoring document variant for each query.

Figure~\ref{fig:supp-interpretation-sample} shows an example based on a region
describing contributions to global growth. The query pair concerning the
largest regional contribution selects $V_2$, whereas the pair concerning the
growth potential of the Asian market selects $V_4$. Thus, variant assignment
remains consistent under meaning-preserving rephrasing within an intent, while
different intents activate different variants. This illustrates that the
learned variants provide stable yet complementary interpretations of the same
document content.